# TokenSTFormer: A Tokenized Spatial-temporal Attention Model for Holistic Motion Analysis in Adolescent Idiopathic Scoliosis Screening

Dong Chen[1], Kenneth M.C. Cheung[1]

[1]The University of Hong Kong
olichen@connect.hku.com; oliver.cd@outlook.com

**Abstract.** Adolescent Idiopathic Scoliosis (AIS) is a prevalent spinal deformity in adolescents that, if left untreated, can result in severe health outcomes. Traditional screening methods are limited by subjective interpretation, reliance on professional expertise and low scalability. To address these challenges, we present ScoliGait dataset, which comprises 1,516 gait video clips paired with corresponding X-ray records. We also introduce TokenSTFormer, a novel model that tokenizes spatial and temporal semantics to enhance feature representation and convergence. Our model achieves state-of-the-art performance, surpassing vanilla Vision Transformer encoder across key metrics, including accuracy of 0.79. This study highlights the potential of leveraging holistic motion features derived from gait video and attention-based models for scalable, cost-effective AIS screening, paving the way for future clinical applications in scoliosis detection.

## 1 Introduction

Adolescent idiopathic scoliosis (AIS) is a structural, lateral curvature of the spine accompanied by vertebral rotation, typically diagnosed during adolescence. Affecting approximate 5% of children globally [1, 2], AIS can lead to severe health consequences if left untreated, such as chronic back pain and psychosocial distress [3, 4]. The current gold standard for diagnosing AIS requires radiographic imaging and measuring the coronal Cobb Angle (CA) >10°, which indicates scoliosis [3]. However, repeated exposure to X-rays raises concerns about cumulative radiation risk, highlighting the need for non-invasive and scalable screening methods.

Early screening is therefore critical to preventing curve progression and enabling timely intervention, especially during adolescence when the spine is still growing. Traditional AIS screening methods, such as the Adams' Forward Bending Test and Scoliometer Measurement, are widely used in clinical and school-based settings [3]. However, their effectiveness and efficiency vary across studies [5, 6], often being influenced by subjective interpretation and factors like participants' obesity [7]. Moreover, these methods face significant challenges in scalability for large-scale screening programs, as they rely on professional expertise, involve high equipment costs, and raise privacy concerns [3].

Advances in technology have introduced alternative approaches for AIS screening, such as using single-camera photographs to analyze back asymmetry [8]. However, methods of using static information fail to incorporate kinematic features providing bio-mechanical insights [9]. Additionally, methods like GaitEdge [10] and SkeletonGait [11] extract spatiotemporal features using synthetic silhouette maps and skeleton maps. Despite their effectiveness, the complex preprocessing pipelines required by these methods limit their practicality in real-world applications.

To address these challenges, we propose ScoliDetect™, a novel system for AIS screening based on gait video analysis, as illustrated in Fig. 1. Specifically, we (1) establish ScoliGait dataset containing 1,516 gait videos paired with corresponding spinal X-ray images and CA measurements. (2) design a de-identified kinematic knowledge map to represent the kinematic features of holistic gait motion, enabling scalable and privacy-conscious screening on mobile devices. (3) introduce TokenSTFormer, a model equipped with Spatial-Temporal Tokenization (STT) to enhance features representation and model convergence.

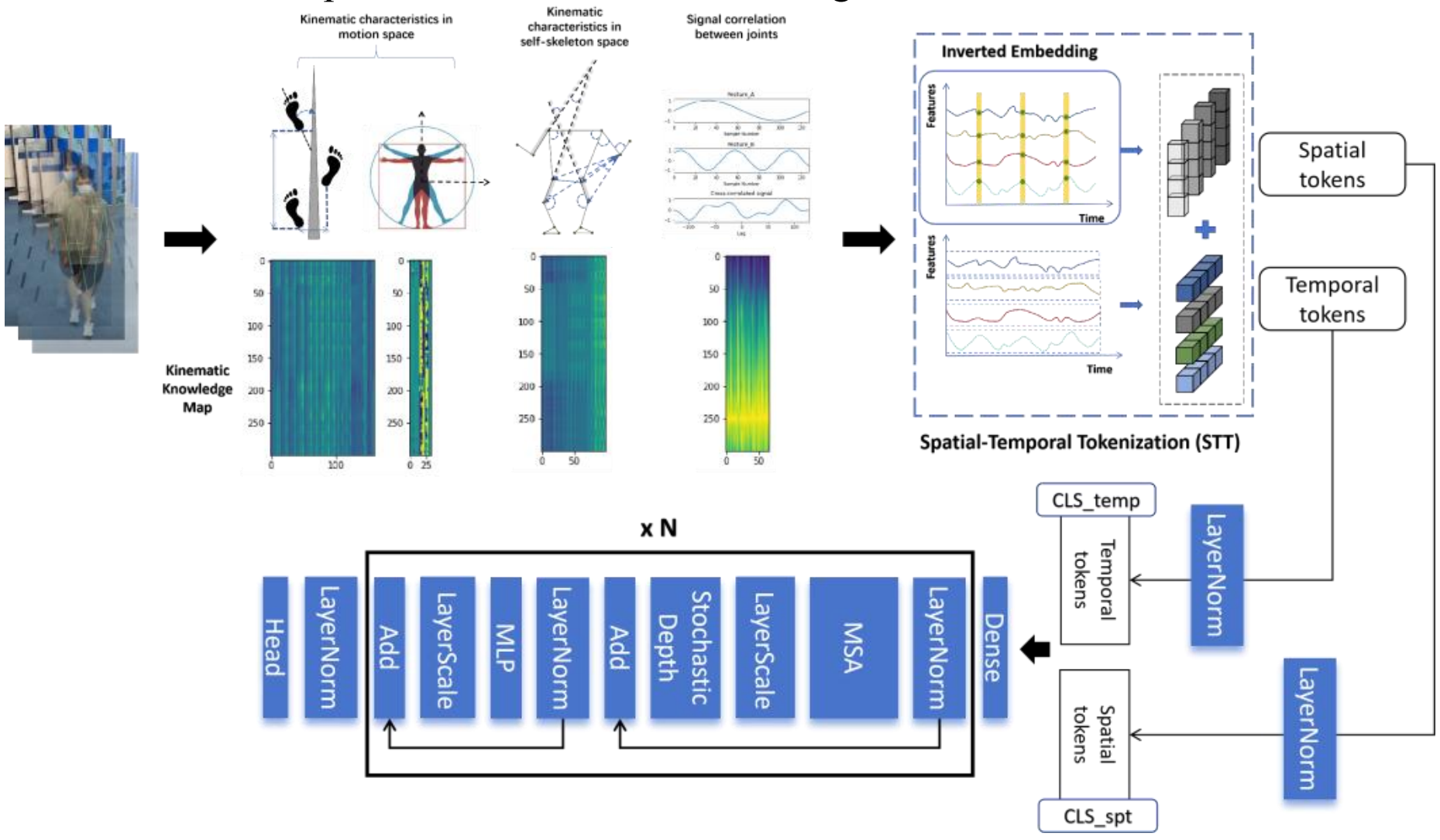

**Fig. 1.** Workflow of ScoliDetect™ system. A gait video recorded using a mobile phone camera is processed to construct a kinematic knowledge map, representing holistic motion features. TokenSTFormer model incorporates Spatial-Temporal Tokenization (STT).

## 2 Dataset

### 2.1 Dataset Description

The ScoliGait dataset was collected using a mobile phone camera at *** Hospital to support scoliosis screening study. A total of 758 participants who signed informed consent form were enrolled in this study. Basic demographic data is summarized in

Table 1. The dataset was expanded by segmenting non-overlapping video clips to enhance inference robustness across different walking periods. Each clip, recorded at 30Hz frame rate and 1080p resolution, captured 5 seconds of holistic walking motion. In total, the final dataset comprises 1,516 video clips.

To the best of our knowledge, ScoliGait is the first dataset to include both gait videos and corresponding spinal X-rays, which serve as the golden standard medical labels. The annotation quality was validated by senior medical doctors. Notably, the ground truth for scoliosis diagnosis relies on radiographic measurements, unlike traditional screening methods which lack sufficient evidence to serve as reliable labels.

**Table 1.** Summary of demographic and clinical attributes in the ScoliGait dataset

| Attributes | Positive (Cobb angle>10°) | Negative (Cobb angle≤10°) |
|---|---|---|
| Number of participants | 758 subjects having informed consent form | |
| Non-overlapped video clips | 1516 non-overlapping clips (150 frames with 30Hz frame rate) | |
| Gender(F/M) | 722/320 | 275/198 |
| Age (mean, std) | 13.86, 2.44 | 11.59, 2.86 |

### 2.2 Data collection and preprocessing

The setup and recording process are illustrated in Fig. 1. The camera was positioned at a height of 2.5 meters to capture shoulder-pelvic angles. Participants were instructed to walk at a natural pace, completing one forward-and-backward cycle along a 4-meter-long path, starting 2 meters away from the camera.

## 3 Methodology

This section involves the way of constructing kinematic knowledge maps (Fig. 1) from pose estimation data and designing STT modules to effectively capture gait features.

Given a video clip

$$V_i^{(t, w, h, c)} = \{f_1, f_2, ..., f_n\} \quad (1)$$

where $f_n$ represents the $n^{th}$ frame of $i^{th}$ subject, (t, w, h, c) represent period, width, height and channel of frames.

$$M_i^{(t, v)} = \emptyset(\tau * F(V_i^{(t, w, h, c)})) \quad (2)$$

where $M_i^{(t, v)}$ represents the kinematic knowledge map with $t$ period and $v$ variates; $F$ is the 2D pose estimation technology; $\tau$ is an amplifying factor which is 1000 in our setting; $\emptyset$ is a prior knowledge function of landmark coordinates transformation.

### 3.1 Kinematic Knowledge Map

To extract kinematic features, we utilized pose estimation technology (YoLoV8) to derive 2D joint coordinates (x, y) from the gait videos [12]. Prior studies have [13, 14] approved that scoliotic gait motion exhibits detectable deviations compared to normal gait. These deviations are primarily induced by musculoskeletal, perceptron or post-adaptive issues. In terms of these prior knowledge, kinematic knowledge map is constructed in three domains: (1) features representing the overall gait pattern in motion space, (2) features capturing the subject's skeletal structure in self-skeleton space, (3) features derived from motion lagging and signal relationships.

The kinematic knowledge map, as shown in Fig. 2, comprises 238 features that represent holistic motion, composing 140 features in motion space, 32 features in self-skeleton space, and 66 features for signal correlation. The numerical values in each section are dependently normalized. Specifically, paired joint-related features are calculated using Euclidean distance, while motion angles between vectors are determined using trigonometric functions. Motion lagging sections are derived through signal cross-correlation, calculated using the SciPy package.

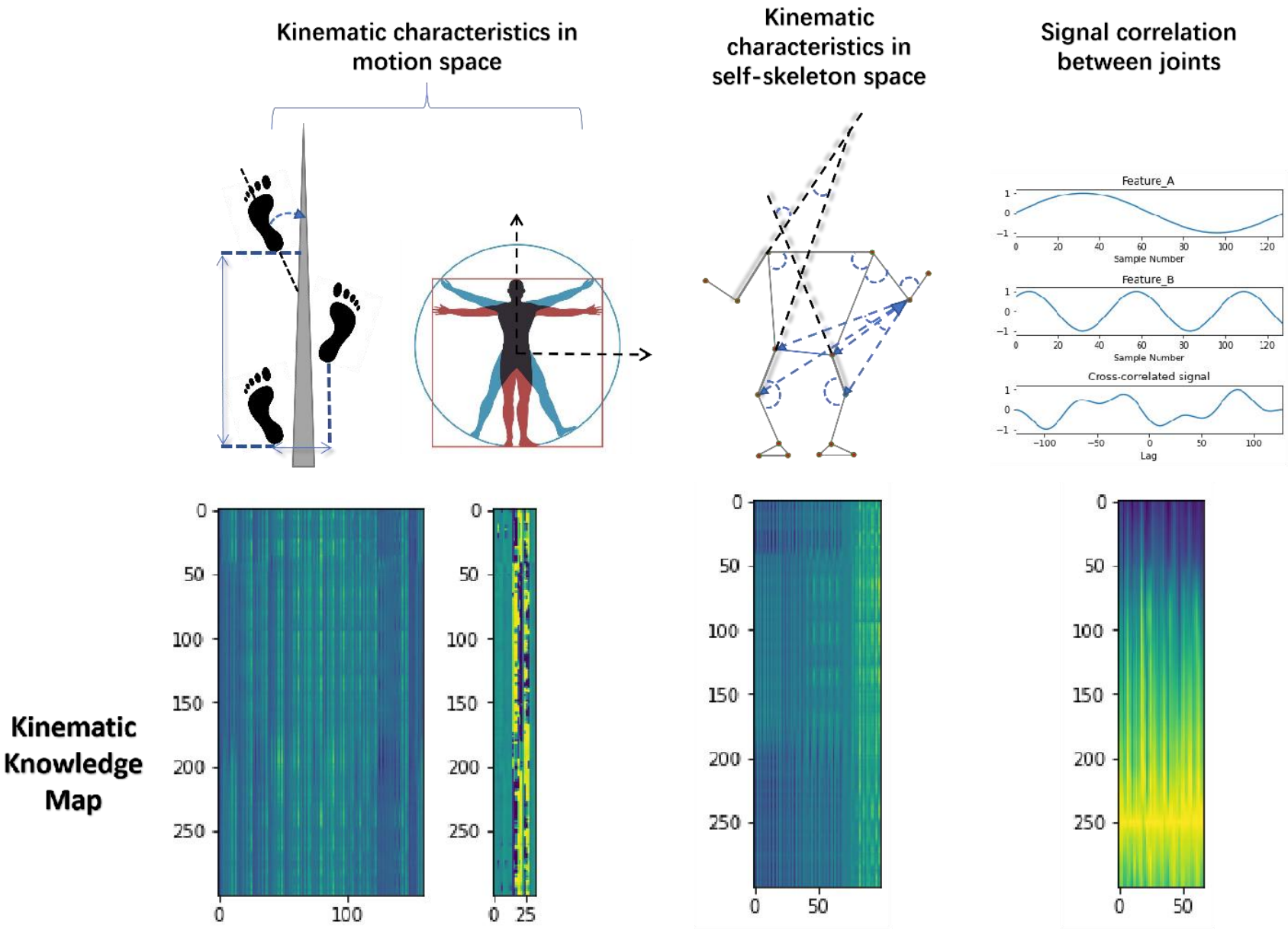

**Fig. 2.** A kinematic knowledge map representing holistic motion features comprises 238 features that represent holistic motion, composing 140 features in motion space, 32 features in self-skeleton space, and 66 features for signal correlation.

### 3.2 Model Architecture

The general architecture of TokenSTFormer is inspired by the Vision Transformer, having residual blocks composed of Multiheaded Self-Attention (MSA) and Multi-Layer Perceptron (MLP) [15]. The proposed Spatial-Temporal Tokenization is illustrated in Fig. 3.

Building on insights from a prior study [16], a Dense layer is applied after spatial tokens to enhance feature representation across variates. Additionally, other key modules, such as LayerScale [17] and Stochastic Depth [18], are integrated in standard configurations.

**Spatial-Temporal Tokenization.** The kinematic knowledge map $M_i^{(t,\ v)}$, with $t$ period and $v$ variates, is tokenized into spatial and temporal tokens using 2D convolutional layers with column-size and row-size kernels, respectively. Temporal tokens $z_{temporal}^{(t,\ d)}$ and spatial tokens $z_{spatial}^{(v,\ d)}$ have $d$ output dimension. Both tokens followed LayerNorm (*LN*) are concatenated as $z_{input}^{(t+v,\ d)}$.

$$z_{input}^{(t+v,\ d)} = concat(LN\left(z_{temporal}^{(t,\ d)}\right), LN(\text{ Dense}(z_{spatial}^{(v,\ d)}))) \tag{3}$$

The main loss function is binary cross entropy (BCM). CLS tokens are respectively applied for temporal and spatial embeddings in our experiments. Auxiliary loss is calculated by Mean Squared Error of these two CLS tokens.

$$Loss_{CLS} = MSE\left(CLS_{temp}, CLS_{spt}\right) \tag{4}$$

$$Loss = Loss_{BCM} + Loss_{CLS} \tag{5}$$

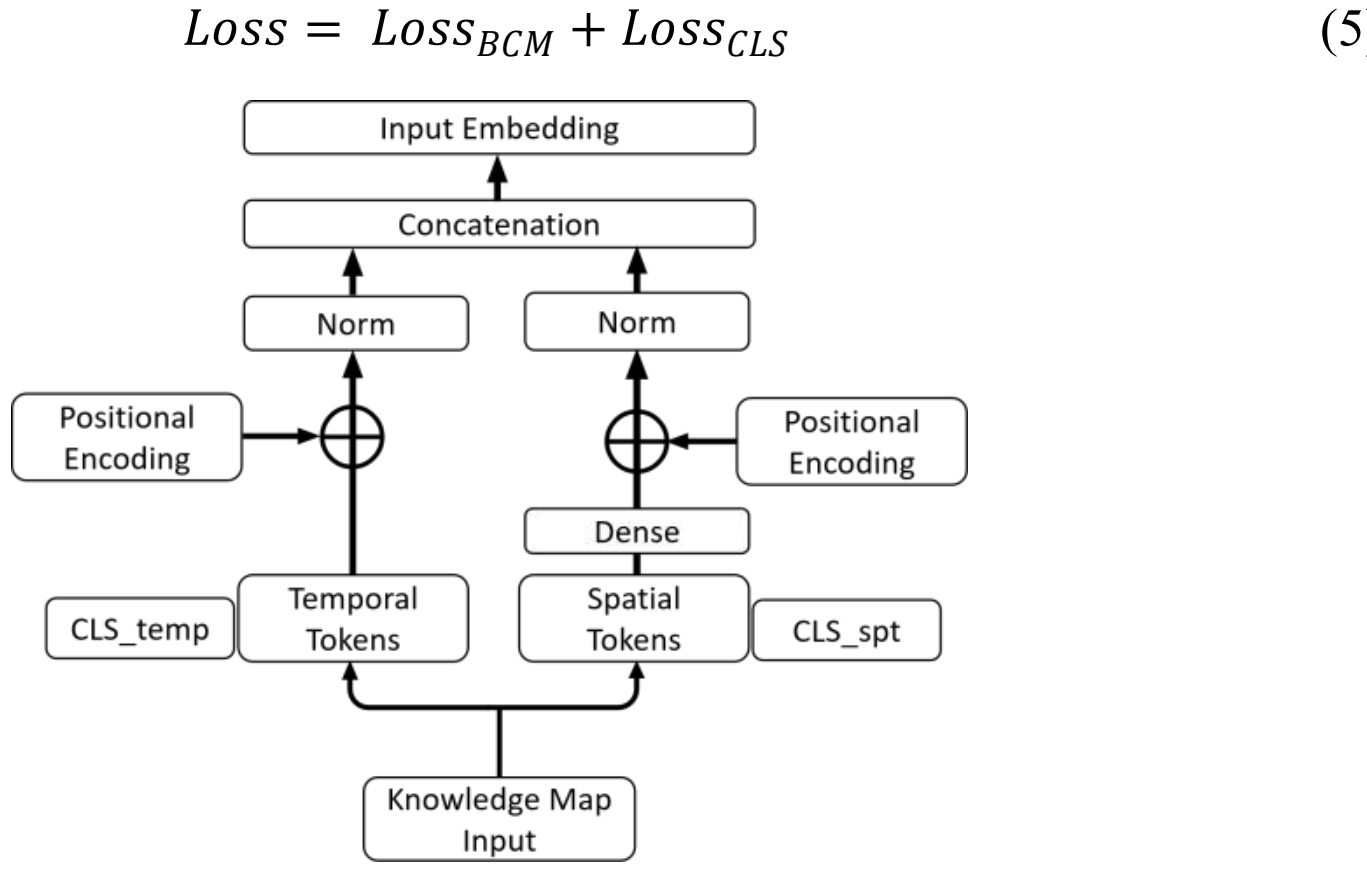

**Fig. 3.** The figure is Spatial-Temporal Tokenization module

## 3.3 Metrics

Key metrics include accuracy, sensitivity, specificity, Positive Predictive Value (PV+), and Negative Predictive Value (PV−). These metrics provide a comprehensive understanding of the model's ability to distinguish between conditions of interest (e.g., disease vs. no disease) based on test outcomes. Here, *TN* represents true negative; *TP* represents true positive; *FN* represents false negative; *FN* represents false negative.

**Accuracy**: Measures the proportion of correctly classified instances (both positive and negative) among all samples.

$$\text{Accuracy} = (\text{TN} + \text{TP})/(\text{TN} + \text{FN} + \text{TP} + \text{FP}) \quad (6)$$

**Positive Predictive Value (PV+)**: Indicates the proportion of true positive predictions among all positive predictions.

$$\text{Positive Predictive Value} = \text{TP}/(\text{TP} + \text{FP}) \quad (7)$$

**Negative Predictive Value (PV−)**: Represents the proportion of true negative predictions among all negative predictions.

$$\text{Negative Predictive Value} = \text{TN}/(\text{TN} + \text{FN}) \quad (8)$$

**Sensitivity**: Measures the ability of the model to correctly identify true positive cases (i.e., the proportion of actual positives correctly predicted).

$$\text{Sensitivity} = \text{TP}/(\text{TP} + \text{FN}) \quad (9)$$

**Specificity**: Reflects the ability of the model to correctly classify true negative cases (i.e., the proportion of actual negatives correctly predicted).

$$\text{Specificity} = \text{TN}/(\text{TN} + \text{FP}) \quad (10)$$

# 4 Experiments

## 4.1 Experimental setup

In this section, the proposed TokenSTFormer model is compared with a vanilla Vision Transformer encoder [15], configured with a 6 by 6 patch size and the same hyperparameters to those of the TokenSTFormer model. The training, validation and testing datasets consisted of 1216, 150, 150 samples, respectively. To reduce class imbalance, we stratified the data samples with a 2.2:1 positive-to-negative ratio. Both categories were adequately represented and minimized bias during model evaluation.

Additionally, the contributions of SST were analyzed in ablation study. Table 2 is the details of hyperparameters setting used in this study.

**Table 2.** Hyperparameter details for model configuration and training

| Model parameters | Training parameters |
|---|---|
| MLP dimension = 384 | Learning rate = 2e-5 |
| Number of heads = 6 | Warmup ratio = 0.1 |
| Number of layers = 5 | Cosine learning rate schedule |
| MHA dimension = 6*256 | Optimizer: Adam |
| Dropout = 0.1 | Batch size = 64 |

### 4.2 Results

Compared to Vision Transformer encoder [15], the proposed TokenSTFormer consistently outperformed the baseline across all evaluation metrics. As shown in Table 3, TokenSTFormer achieved accuracy of 0.787, demonstrating its superior overall classification capability. Additionally, it achieved a sensitivity of 0.845 and specificity of 0.660, underscoring its effectiveness in correctly identifying both positive and negative cases.
Moreover, TokenSTFormer excelled in predictive values, with PV+ (0.845) and PV− (0.660), highlighting its reliability in making accurate and balanced predictions. These results highlighted the robustness and efficiency of TokenSTFormer for AIS screening.

**Table 3.** Comparison of evaluation metrics among TokenSTFormer, Vision Transformer encoder, and traditional methods

| | Accuracy | Sensitivity | Specificity | PV+ | PV- |
|---|---|---|---|---|---|
| | | 0.46 | 0.84 | 0.30 | |
| Paper Report [5, 6] | | 0.51 | 0.96 | 0.95 | 0.53 |
| | | 0.37 | 0.90 | 0.80 | 0.59 |
| Average | | 0.447 | 0.900 | 0.683 | 0.560 |
| Transformer encoder | 0.740 | 0.796 | 0.617 | 0.820 | 0.580 |
| TokenSTFormer | **0.787** | **0.845** | **0.660** | **0.845** | **0.660** |

### 4.3 Ablation studies

We conducted ablation studies to analyze certain feature effects on performance. As shown in Table 4, LayerNorm and independent positional encoding play critical roles in SST modules to ensure the model's accuracy and robustness.

**Table 4.** Metrics comparison among TokenSTFormer, w/o Spatial-Temporal Tokenization.

| | Accuracy | Sensitivity | Specificity | PV+ | PV- |
|---|---|---|---|---|---|
| SST w/o LayerNorm | 0.720 | 0.738 | 0.681 | 0.835 | 0.542 |
| Single pos encoding | 0.687 | 0.728 | 0.600 | 0.798 | 0.500 |
| TokenSTFormer | **0.787** | **0.845** | **0.660** | **0.845** | **0.660** |

**Spatial-Temporal Tokenization.** The purpose of STT is to separate and normalize temporal and spatial tokens, thereby minimizing the distance between two types of tokens. To further analyze its impact, we analyzed the cosine similarity of spatial-temporal tokens between baseline model and that without LayerNorm in STT, as illustrated in Fig. 4. The majority of points lie above the red dashed line (slope = 1), indicating that the cosine similarity in the baseline model is significantly smaller. These results demonstrate that STT effectively learns temporal-spatial specific transformations to reproject tokens into an enhanced semantic space.

**Number of layers.** We evaluated the performance metrics of models with varying numbers of attention blocks. As illustrated in Fig. 5, the model achieves the highest accuracy when the number of attention blocks is set to 5. This suggests that an optimal balance between model complexity and performance is achieved at this configuration.

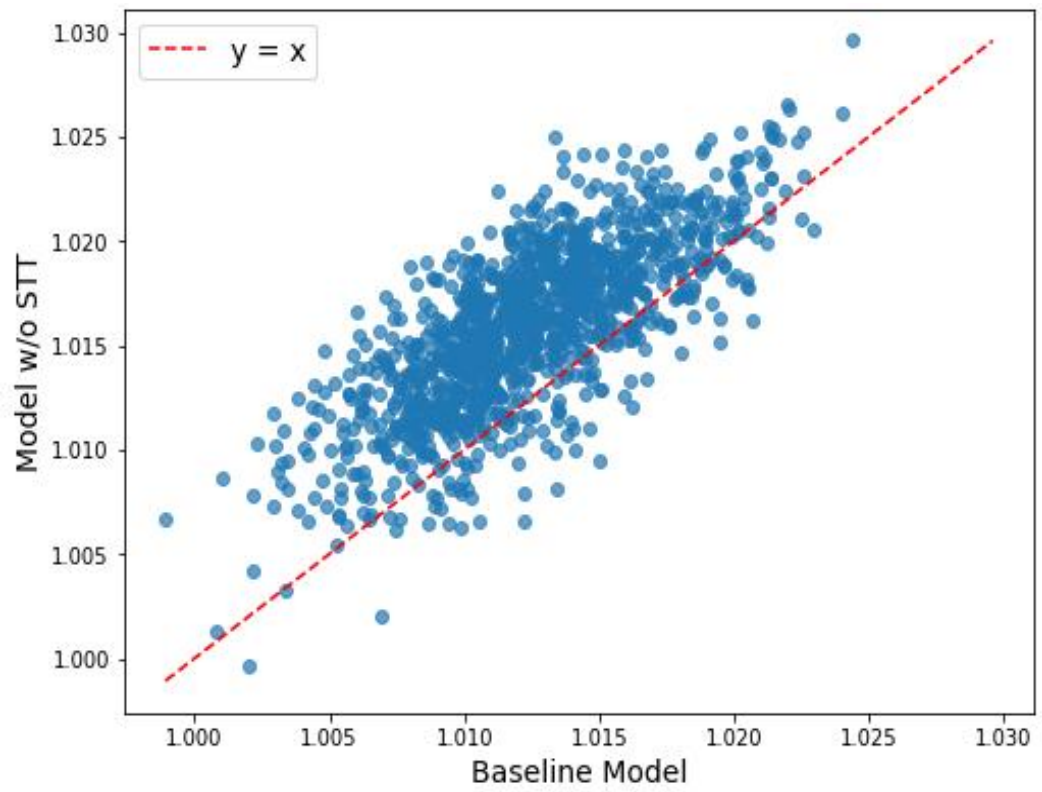

**Fig. 4.** Scatter plot comparing token cosine similarity distances between the TokenSTFormer and that without STT.

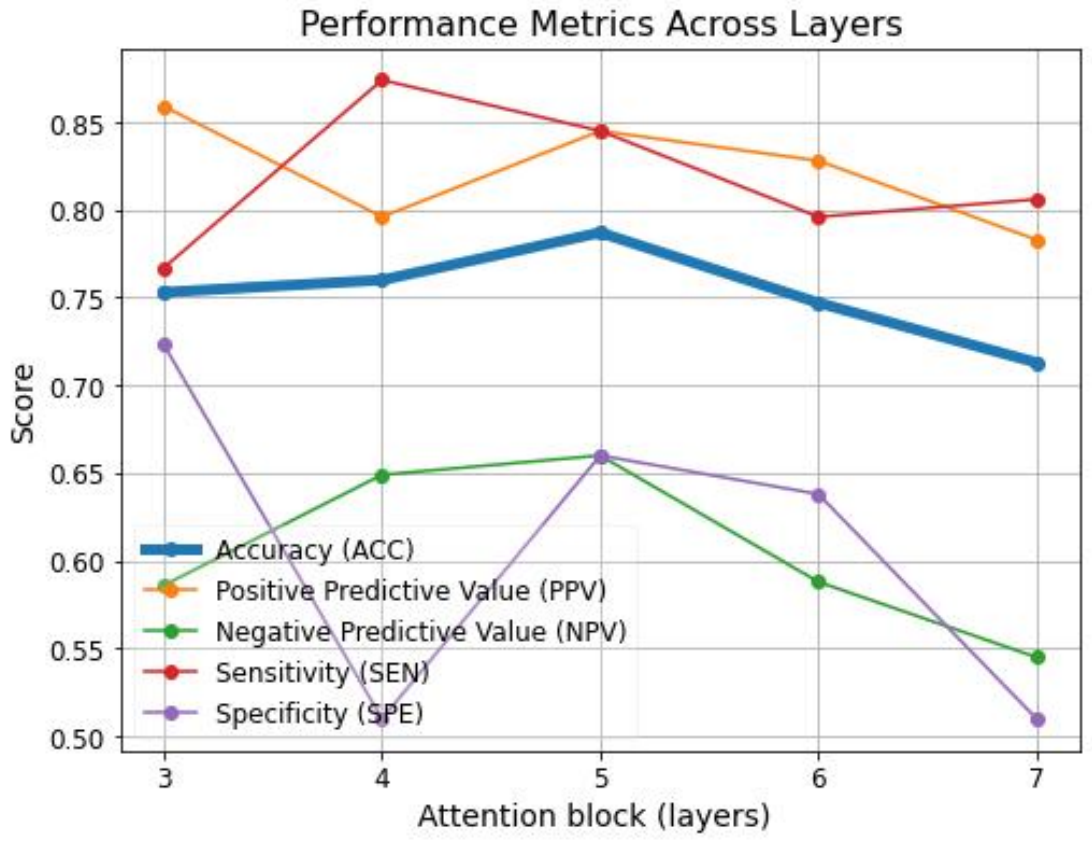

**Fig. 5.** The plot shows the variation in performance as the number of attention block (layers) increases.

# 5 Conclusions

In this study, we introduced ScoliDetect$^{TM}$, an AI-assisted holistic motion analysis system utilizing a smartphone camera for scalable and cost-effective scoliosis screening. By leveraging Spatial-Temporal Tokenization (SST), the proposed TokenSTFormer effectively learns robust spatial-temporal features. This work represents a promising step toward the deployment of accessible, accurate, and efficient diagnostic tools for scoliosis screening and monitoring.